\pdfoutput=1
\documentclass[11pt]{article}

\usepackage[]{emnlp2021}

\usepackage{times}
\usepackage{latexsym}

\usepackage{graphicx}
\usepackage{url}
\usepackage{multirow}
\usepackage[ruled,noend]{algorithm2e}
\SetKwInput{KwInput}{Input}
\SetKwInput{KwOutput}{Output}
\SetEndCharOfAlgoLine{}
\usepackage{algpseudocode}
\usepackage{mathtools}
\usepackage{amsfonts}
\usepackage{xcolor}
\usepackage{enumitem}
\usepackage{marvosym}
\usepackage{textcomp}
\usepackage{array}
\usepackage{graphicx}
\usepackage{url}
\usepackage{pifont}
\usepackage{amsmath}
\usepackage{booktabs}
\usepackage{makecell}
\usepackage{subfigure}

\newcommand\blfootnote[1]{%
  \begingroup
  \renewcommand\thefootnote{}\footnote{#1}%
  \addtocounter{footnote}{-1}%
  \endgroup
}

\usepackage[T1]{fontenc}
\usepackage[utf8]{inputenc}

\usepackage{microtype}

\title{Uncovering Main Causalities for Long-tailed Information Extraction}

\author{Guoshun Nan$^{1}$\thanks{$^{*}$Contributed equally.}\;, Jiaqi Zeng$^{2*}$, Rui Qiao$^{3*}$, Zhijiang Guo$^{4}$ \and Wei Lu$^{1}$\\
$^{1}$StatNLP Research Group, Singapore University of Technology and Design \\
  $^{2}$Carnegie Mellon University \ \ 
  $^{3}$National University of Singapore \ \ 
  $^{4}$University of Cambridge \\
  \texttt{nanguoshun@gmail.com, jiaqizen@andrew.cmu.edu} \\
  \texttt{qiaorui@comp.nus.edu.sg, zg283@cam.ac.uk}\\
  \texttt{luwei@sutd.edu.sg}\\}

\begin{document}
\maketitle
\begin{abstract}
Information Extraction (IE)\blfootnote{Accepted as a long paper in the main conference of EMNLP 2021 (Conference
on Empirical Methods in Natural Language Processing).} aims to extract structural information from unstructured texts. In practice, long-tailed distributions caused by the selection bias of a dataset, may lead to {\color{black}incorrect} correlations,  {\color{black}also known as spurious correlations,} between entities and labels in the conventional likelihood models. This motivates us to propose counterfactual IE (CFIE), a novel framework that aims to uncover the main causalities behind data in the view of causal inference. Specifically, 1) we first introduce a unified structural causal model (SCM) for various IE tasks, describing the relationships among variables; 2) with our SCM, we then generate counterfactuals based on an explicit language structure to better calculate the direct causal effect during the inference stage; 3) we further propose a novel debiasing approach to yield more robust predictions. Experiments on three IE tasks across five public datasets show the effectiveness of our CFIE model in mitigating the spurious correlation issues. %Code will be released upon acceptance.
\end{abstract}

%Previous efforts towards unbiased predictions may be unsatisfactory for the long-tailed IE, as language structures that have been proven crucial for the IE tasks, are largely ignored. 

\section{Introduction}
\begin{figure}
  \begin{center}
    \includegraphics[width=0.44\textwidth]{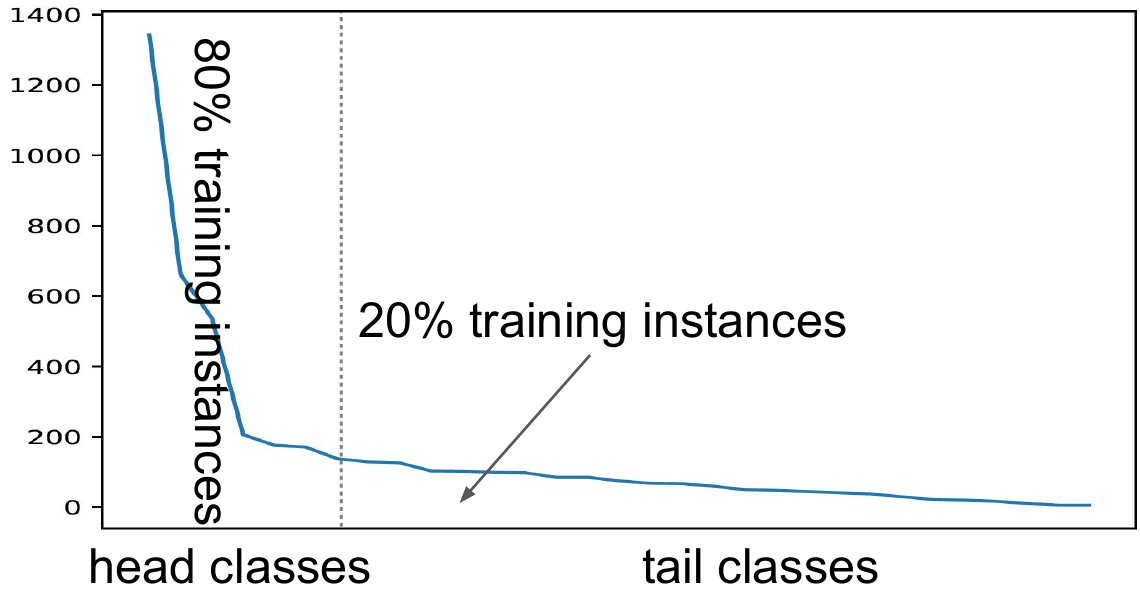}
  \end{center}
%  \vspace{-4mm}
  \caption{Class distribution of the ACE2005 dataset. }
  \label{fig:intro}
%  \vspace{-3mm}
\end{figure}

The goal of Information Extraction (IE) is to detect the structured information from unstructured texts. Previous deep learning models for IE tasks, such as named entity recognition (NER; \citealt{lampleetal2016}), relation extraction (RE; \citealt{ Peng2017CrossSentenceNR}) and event detection (ED; \citealt{nguyengrishman2015event}), {\color{black} are largely proposed for learning under some reasonably balanced label distributions. However, in practice, these labels usually follow a long-tailed distribution~\citep{doddington2004automatic}.} Figure~\ref{fig:intro} shows such an unbalanced distribution on the ACE2005~\citep{doddington2004automatic} dataset. As a result, performance on the instance-scarce (tail) classes may drop significantly.
For example, on {\color{black} an existing model} for NER~\citep{jie2019dependency}, the macro F1 score of instance-rich (head) classes can be 71.6, while the score of tail classes sharply decreases to 41.7.

%independent and identically distributed (i.i.d.) data.
%on OntoNotes5.0 ~\citep{pradhan2013} for NER, and 

The underlying causes for the above issue are the biased statistical dependencies between {\color{black} entities}\footnote{For the NER task, a model may also build the spurious correlations between the part-of-speech (POS) tags of entities and class labels. For RE and ED tasks, a model may also learn incorrect correlations between features like NER tags and labels. We consider all above issues in our proposed causal diagram.} and classes, known as spurious correlations~\citep{srivastava2020robustness}. For example, an entity \textit{Gardens} appears 13 times in the training set of OntoNotes5.0, with the NER tag location  \textit{LOC}, and only 2 times as organization \textit{ORG}. A classifier trained on this dataset tends to build spurious correlations between \textit{Gardens} and \textit{LOC}, although \textit{Gardens} itself does not indicate a location. Most existing works on addressing spurious correlations focus on images, such as re-balanced training~\citep{lin2017focal}, transfer learning~\citep{liu2019large} and decoupling~\citep{kang2019decoupling}. However, these approaches may not be suitable for natural language inputs. Recent efforts on information extraction~\citep{han2018hierarchical, zhangetal2019long} incorporate prior knowledge, which requires data-specific designs.

% To address this issue, previous efforts on information extraction~\citep{han2018hierarchical, zhangetal2019long} utilize data-specific prior information, which may not be suitable for other datasets. Other solutions mainly proposed for visual tasks, including re-balanced training \citep{lin2017focal}, transfer learning \citep{liu2019large}, and decoupling \citep{kang2019decoupling} strategy, may still suffer from the under-fitting (over-fitting) issue for head (tail) classes \citep{tang2020longtailed}. 

% One may also think that BERT-based models \citep{devlin2019bert} may address the long-tailed issues. However, we empirically show that such models still have limitations.

Causal inference~\citep{pearl2016causal} is promising in tackling the above spurious correlation issues caused by unbalanced data distribution. Along this line, various causal models have been proposed for visual tasks~\citep{abbasnejad2020counterfactual,tang2020unbiased}. Despite their success, these methods may be unsatisfactory on textual inputs. Unlike images, which can be easily disentangled with detection or segmentation methods for causal manipulation, texts rely more on the context involving complex syntactic and semantic structures. Hence it is impractical to apply the methods used in images to disentangle tokens' representations. Recent causal models~\citep{zeng2020counterfactual,wang2020identifying,wang2020robustness} on text classification eliminate biases by replacing target entities or antonyms. These methods do not consider structural information, which has proven effective for various IE tasks as they are able to capture non-local interactions~\citep{zhang2018graph, jie2019dependency}. This motivates us to propose a novel framework termed as {\color{black}counterfactual information extraction (CFIE)}. Different from previous efforts, our CFIE model alleviates the spurious correlations by generating counterfactuals~\citep{bottou2013counterfactual, abbasnejad2020counterfactual} based on the syntactic structure~\citep{zhang2018graph}.

% Recently proposed models \cite{zeng2020counterfactual,wang2020identifying,wang2020robustness} generate counterfactuals by replacing target entities or antonyms, while the 
% dependency structure that has been proved to be effective for the IE tasks \cite{zhang2018graph, jie2019dependency}, is still under-explored. For different IE tasks, the way of utilizing the syntax structure varies: the RE task may rely more on the context and entity type rather than entities themselves, while classifications in NER and ED tasks count more on entities than the context. Hence, it is challenging to decide properly on how to utilize language structures for different IE tasks, in the view of causal inference.  

% The above discussions motivate us to propose CFIE, a novel framework that combines the syntax structure and counterfactual analysis in causal inference \citep{pearl2016causal} to alleviate the spurious correlations of the long-tailed IE tasks including NER, RE and ED.

From a causal perspective, counterfactuals state the results of the outcome if certain factors had been different. This concept entails a hypothetical scenario where the values in the causal graph can be altered to study the effect of the factor. Intuitively, the factor that yields the most significant changes in model predictions has the greatest impact and is therefore considered as the main effect. Other factors with minor changes are categorized as side effects. In the context of IE with language structures, counterfactual analysis answers the question on ``which tokens in the text would be the key clues for RE, NER or ED that could change the prediction result''. With that in mind, our CFIE model is proposed to explore the language structure to eliminate the bias caused by the side effects and maintain the main effect for prediction. We show the effectiveness of CFIE on three representative IE tasks including NER, RE and ED. {\color{black}Our code} and the supplementary materials are available at \url{https://github.com/HeyyyyyyG/CFIE}.

Specifically, our major contributions are:

\begin{itemize}
    \item To the best of our knowledge, CFIE is the first study that marries the counterfactual analysis and syntactic structure to address the spurious correlation issue for long-tailed IE. We build different structural causal models (SCM; \citealt{pearl2016causal}) for various IE tasks to better capture the underlying main causalities.
    \item To alleviate spurious corrections, we generate counterfactuals based on syntactic structures. To achieve more robust predictions, we further propose a novel debiasing approach, which maintains a better balance between the direct effect and counterfactual representations.
    \item Extensive quantitative and qualitative experiments on various IE tasks across five datasets show the effectiveness of our approach.
\end{itemize}

\section{Model} \label{model}
\begin{figure*}
    \centering
    \includegraphics[scale=0.5]{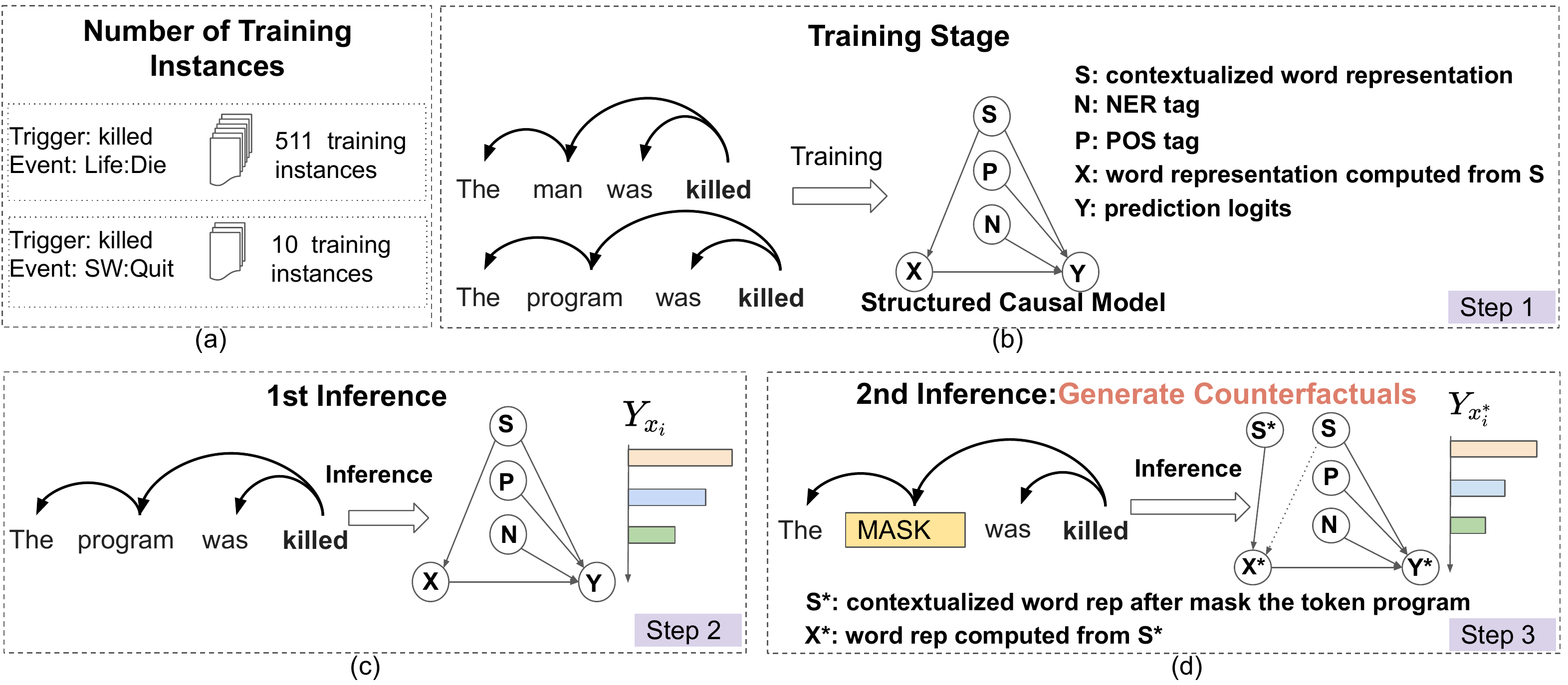}
%    \vspace{-1mm}
    \caption{Training and inference stages of CFIE for ED, which detects the triggers and then predict the corresponding event types for a given sentence. (a) The number of training instances for the trigger \textit{killed} labeled with two different event types. (b) Step $1$ builds the SCM and trains the model. (c) Step $2$ obtains prediction results for each token, e.g., $Y_{x_i}$ for the token \textit{killed}. (d) Step $3$ generates the counterfactuals of each token by masking the tokens along the 1st hop of a syntactic tree, and then achieves counterfactual prediction, e.g., $Y_{x^*_i}$ for \textit{killed}. }%, which consists of five steps to improve classifications for instance-scarce classes.
    \label{fig:arc1}
%    \vspace{-2mm}
\end{figure*}

Figure~\ref{fig:arc1} demonstrates the proposed CFIE method using an example from the ACE2005 dataset ~\citep{doddington2004automatic} on the ED task. % which consists of three key steps, i.e., causal representation learning, counterfacutals generation and causal effect estimation.%, to address the long-tailed issues in the information extraction. 
As shown in Figure~\ref{fig:arc1} (a), two event types ``\textit{Life:Die}'' and ``\textit{SW:Quit}'' of the trigger \textit{killed} have $511$ and $19$ training instances, respectively. Such an unbalanced distribution may mislead a model to build spurious correlations between the trigger word \textit{killed} and the type ``\textit{Life:Die}''. The goal of CFIE is to alleviate such incorrect correlations. CFIE employs SCM~\citep{pearl2016causal} as causal diagram as it clearly describes relationships among variables. We give the formulation of SCM as follows. 

\paragraph{SCM:} Without loss of generality, we express SCM as a directed acyclic graph (DAG) $\mathcal{G} = \{\mathbb{V}, \mathbb{F}, \mathbf{U} \}$, where the set of observables (vertices) are denoted as $\mathbb{V} = \{V_1, ...,V_n\}$, the set of functions (directed edges) as $\mathbb{F} = \{f_1,..., f_n\}$, and the set of exogenous variables (e.g., noise) \cite{pearl2016causal} as $\mathbf{U} = \{U_1, ..., U_n\}$ for each vertice. Here $n$ is the number of nodes in $\mathcal{G}$. In the deterministic case where $\mathbf{U}$ is given, the values of all variables in SCM are uniquely determined. Each observable $V_i$ can be derived from:
\begin{equation}
V_i :=  f_i(\mathbf{PA}_i, U_i), (i=1,...,n),   
\end{equation}
where $\mathbf{PA}_i \subseteq \mathbf{V} \backslash V_i$ is the set of parents of $V_i$ and ``$\backslash$'' is an operator that excludes $V_i$ from $\mathbf{V}$, and $f_i$ refers to the direct causation from $\mathbf{PA}_i$ to its child variable $V_i$. Next we show how our SCM-based CFIE works.

%\forall i$,

\begin{figure}
  \begin{center}
    \includegraphics[width=0.45\textwidth]{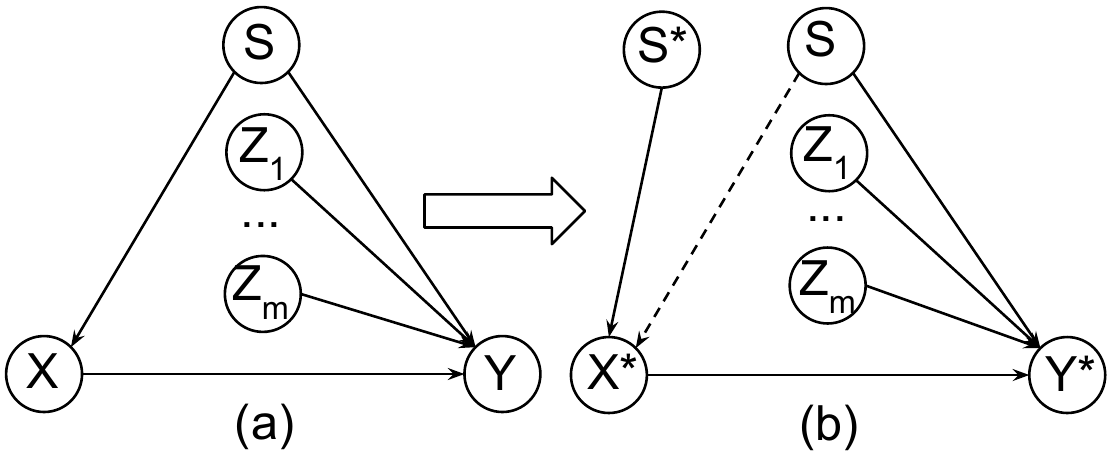}
  \end{center}
%  \vspace{-4mm}
  \caption{(a) a unified structured causal model (SCM) for IE tasks. (b) causal interventions on $X$.}
  \label{fig:scm}
%  \vspace{-4mm}
\end{figure}

\subsection{Causal Representation Learning} \label{step1}
Figure~\ref{fig:scm} presents our unified SCM $\mathcal{G}_{ie}$ for IE tasks based on our prior knowledge. The variable $S$ indicates the contextualized representations of an input sentence, where the representations are the output from a BiLSTM~\citep{Schuster1997BidirectionalRN} or a pre-trained BERT encoder~\citep{devlin2019bert}. $Z_j$ ($j \in \{1,2,\dots, m\}$) represents features such as the NER tags and part-of-speech (POS) tags, where $m$ is the number of features. The variable $X$ is the representation of a relation, an entity and a trigger for RE, NER and ED, respectively, and $Y$ indicates the output logits for classification. 

\iffalse
The two well-known causal inference frameworks are SCMs and potential outcomes~\citep{rubin2019essential} which are complementary and theoretically connected. We choose SCMs in our case due to their advantages in expressing and reasoning about the effects of causal relationships among variables.
\fi

% \noindent
% \textbf{Structural Causal Models (SCM):}  
%, including  $X$ $\rightarrow$ $Y$, $S$ $\rightarrow$ $Y$, $Z_1$ $\rightarrow$ $Y$, $\dots$, $Z_m$ $\rightarrow$ $Y$

For $\mathcal{G}_{ie}$, we denote the parents of $Y$ as $\mathcal{E} = \{S,X, Z_1,\dots, Z_m \}$. The direct causal effects towards $Y$ are linear transformations. Transformation for each edge $i \rightarrow Y$ is denoted as $\mathbf{W}_{iY} \in \mathbb{R}^{c \times d }$, where $i \in \mathcal{E}$, $c$ is the number of classes, and $d$ is the dimensional size. We let $\mathbf{H}_i \in \mathbb{R}^{d \times k}$ denote $k$ representations for the node $i$. Then, the prediction can be obtained by summation $Y_x = \sum_{i \in \mathcal{E}} \mathbf{W}_{iY} \mathbf{H}_i$ or gated mechanism $Y_x = \mathbf{W}_g \mathbf{H}_X \odot \sigma (\sum_{i \in \mathcal{E}} \mathbf{W}_{iY} \mathbf{H}_i)$, where $\odot$ refers to element-wise product, $\mathbf{H}_X$ is the representation of the node $X$, and $\mathbf{W}_g \in \mathbb{R}^{c \times d}$ and $\sigma(\cdot)$ indicate a linear transformation and the sigmoid function, respectively. 

To avoid any single edge dominating the generation of the logits $Y_x$, we introduce a cross-entropy loss $\mathcal{L}_{iY}, i \in \mathcal{E}$ for each edge. Let $\mathcal{L}_Y$ denote the loss for $Y_x$, the overall loss $\mathcal{L}$ can be:
\begin{equation}
  \label{eq:loss}
   \mathcal{L} = \mathcal{L}_{Y} + \sum_{i \in \mathcal{E}} \mathcal{L}_{iY} 
\end{equation}

%\footnote{The mathematical symbol $k \geq 2$ is the length of sequence for NER and ED, and $k = 1$ for RE.}
Step $1$ in Figure~\ref{fig:arc1} (b) trains the above causal model, aiming to teach the model to identify the main cause (main effect) and the spurious correlations (side effect) for classification. Our proposed SCM is encoder neutral and it can be equipped with various encoders, such as BiLSTM and BERT. %For simplicity, we omit exogenous variables $\mathbf{U}$ from the graph as these variable are only useful for the derivations in the following sections. 

% \noindent
% \textbf{Fusing Syntactic Structures Into SCM:} %Syntactical structure have been proven useful for IE tasks as it is able to capture underlying rules and patterns of sentences. 

\paragraph{Fusing Syntactic Structures Into SCM:} So far we have built our unified SCM for IE tasks. On the edge $S$ $\rightarrow$ $X$, we adopt different neural network architectures for RE, NER and ED. For RE, we use dependency trees to aggregate long-range relations with graph convolution networks (GCN; \citealt{Kipf2016SemiSupervisedCW}) to obtain $\mathbf{H}_X$. For NER and ED, we adopt the dependency-guided concatenation approach~\citep{jie2019dependency} to obtain $\mathbf{H}_X$. %Based on the proposed SCM, we train the model to learn better causal representations.

\subsection{Inference and Counterfactual Generation}
Given the above SCM, we train our neural model designed for a specific task such as ED. Step $2$ in Figure \ref{fig:arc1} (c) performs inference with our proposed SCM, and Step $3$ in Figure \ref{fig:arc1}(d) generates syntax-based counterfactuals to obtain better main effect.  
%$\hat{f}_i(\mathbf{\hat{PA}}_i, U_i)$

% \noindent
% \textbf{Interventions:}

\paragraph{Interventions:} For $\mathcal{G}_{ie}$, an intervention indicates an operation that modifies a subset of variables $\mathbf{V} \subseteq \mathbb{V}$ to new values where each variable $V_i \in \mathbf{V} $ is generated {\color{black} by manual manipulations}. Thus, the causal dependency between $V_i$ and its parents $\{\mathbf{PA_i}, U_i\}$ 
will be cut off, as shown in Figure~\ref{fig:scm}(b). Such an intervention for one variable $X \in \mathbb{V}$ can be expressed by the $do$-notation $do(X = x^*)$ where $x^*$ is the given value {\color{black}\cite{pearl2009causality}}.

%Recall from Section \ref{model} the definition of SCM and the set of exogenous variables $U$ with value $u$.
\paragraph{Counterfactuals:} Unlike interventions, the concept of counterfactual reflects an imaginary scenario for ``what would the outcome be had the variable(s) been different''. Let $Y \in \mathbb{V}$ denote the outcome variable, and let $X \in \mathbb{V} \backslash \{Y\}$ denote the variable of study. The counterfactual is obtained by setting $X=x^*$ and formally estimated as:
 \begin{equation}
Y_{x^*}(u) = Y_{\mathcal{G}_{x^*}}(u)
 \end{equation}
where $\mathcal{G}_{x^*}$ means assigning $X=x^*$ for all equations in the SCM $\mathcal{G}$. We slightly abuse the notation and use $Y_{x^*}$ {\color{black}as a short form} for $Y_{x^*}(u)$, since the exogenous variable $u$ is not explicitly required here\footnote{Derivations are given in the supplementary materials.}. For SCM $\mathcal{G}$, the counterfactual $Y_{x^*}$ of the original instance-level prediction $Y_x$ is computed as:
 \begin{equation}
 \begin{aligned}
 	Y_{x^*} &= f_Y(do(X=x^*), S=s, Z=z) \\
 	&= \sum_{\substack{i \in \mathcal{E} \backslash \{X\}}} \mathbf{W}_{iY} \mathbf{H}_i + \mathbf{W}_{XY} \mathbf{H}_{x^*} 
 \end{aligned}
 \end{equation}
where $f_Y$ is the function that computes $Y$. Compared to the vanilla formula for $Y_x$, we only replace its feature $\mathbf{H}_X$ with $\mathbf{H}_{x^*}$.

\begin{figure}
  \begin{center}
    \includegraphics[width=0.48\textwidth]{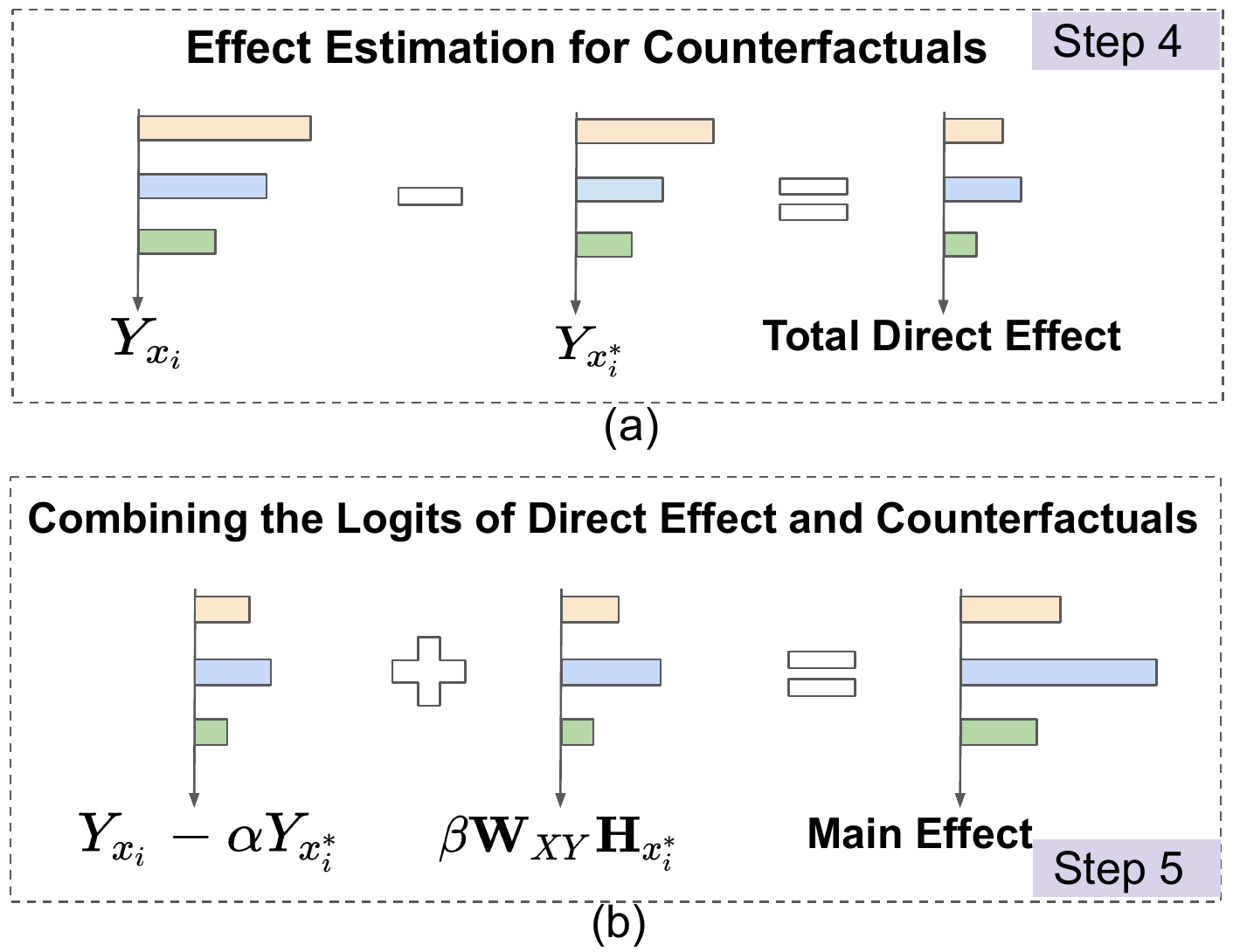}
  \end{center}
%  \vspace{-4mm}
  \caption{Causal effect estimation: (a) Step $4$ computes the TDE by subtraction of outputs for each token in Step $2$ and $3$, e.g., $Y_{x_i} - \alpha Y_{x^*_i}$. (b) Step $5$ obtains more robust predictions by highlighting each token's counterfactual representations, e.g., {\color{black} $ \mathbf{W}_{XY} \mathbf{H}_{x^*_i}$} for \textit{killed}.}
  \label{fig:arc2}
%  \vspace{-4mm}
\end{figure}
\paragraph{Counterfactual Generation:} There are many language structures such as dependency and constituency {\color{black}trees}~\citep{Marcus1993BuildingAL}, semantic role labels~\citep{Palmer2005ThePB}, and abstract meaning {\color{black} representations}~\citep{banarescuetal2013abstract}. We choose the dependency tree in our case as it can capture rich relational information and complex long-distance interactions that have proven effective on IE tasks. Counterfactuals lead us to think about: what are the key clues that determine the relations of two entities for RE, and a certain span of a sentence to be an entity or an event trigger for NER and ED respectively? {\color{black} As demonstrated in Figure \ref{fig:arc1} (d)}, we mask entities, or the tokens in the scope of 1 hop on the dependency tree. Then this masked sequence is fed to a BiLSTM or BERT encoder to output new contextualized representations $S^*$, as shown in Figure \ref{fig:arc1} (d). Then we feed $S^*$ to {\color{black} the function of the edge $S \rightarrow X$} to get $X^*$. This operation also aligns with a recent finding \citep{zeng2020counterfactual} that the entity itself may be more important than context in NER. By doing so, the key clues are expected to be wiped off in {\color{black}the representations $X^*$ of counterfactuals}, strengthening the main effect while reducing the spurious correlations.% and the side effect.  

\subsection{Causal Effect Estimation}
As shown in Figure \ref{fig:arc2}, we estimate the causal effect in the Step $4$ and use the representation of counterfactuals for a more robust prediction in Step $5$. Inspired by SGG-TDE \citep{tang2020unbiased}, we compare the original outcome $Y_x$ and its counterfactual $Y_{x^*}$ to estimate the main effect so that the side effect can be alleviated with Total Direct Effect (TDE) \cite{pearl2009causality}\footnote{Derivations are given in the supplementary materials.}:
\begin{equation}
{\rm TDE} = Y_x - Y_{x^*} 
\label{eq:tde}
\end{equation}

As both of the context and entity (or trigger) play important roles for the classification in {\color{black}NER, ED, and RE}, we propose a novel approach to further alleviate the spurious correlations caused by side effects, while strengthening the main effect at the same time. The interventional causal effect of the $i$-th entity in a sequence can be described as:
\begin{equation}
    {\rm Main \ Effect} = Y_{x_i} - \alpha Y_{x_i^*} + \beta \mathbf{W}_{XY} \mathbf{H}_{x^*_i}
    \label{eq:debias}
\end{equation}
where $\alpha, \beta$ are the hyperparameters that balance the importance of context and entity (or trigger) for NER, ED, and RE. The first part $Y_{x_i} - \alpha Y_{x_i^*}$ indicates the main effect, which reflects more about the debiased context, while the second part $\mathbf{W}_{XY} \mathbf{H}_{x^*_i}$ reflects more about the entity (or trigger) itself. Combining them yields more robust prediction by better distinguishing the main and side effect. 

As shown in Step 4 of Figure~\ref{fig:arc2}(a), the sentence ``The program was killed'' produces a biased high score for the event ``\textit{Life:Die}'' in $Y_x$ and results in wrong prediction due to the word ``\textit{killed}''. By computing the counterfactual $Y_{x^*}$ with ``program'' masked, the score for ``\textit{Life:Die}'' remains high but the score for ``\textit{SW:Quit}'' drops. The difference computed by $Y_{x} - \alpha Y_{x^*}$ {\color{black}may help us to correct the prediction while understanding the important role of the word ``program''.} However, we may not only rely on the context since the entity (trigger) itself is also an important clue. To magnify the difference and obtain more robust predictions, we strengthen the impact of entity (trigger) on the final results by $\mathbf{W}_{XY} \mathbf{H}_{x^*_i}$ as shown in Step 5 of Figure \ref{fig:arc2}(b). Such a design differs from SGG-TDE \citep{tang2020longtailed} by providing more flexible adjustment and effect estimation {\color{black}with hyperparameters $\alpha$ and $\beta$}. We will show that our approach is more suitable for long-tailed IE tasks in experiments.  

\section{Experiments}
\subsection{Datasets and Settings}
We use five datasets in our experiments including OntoNotes5.0~\citep{pradhan2013} and ATIS~\citep{tur2010left} for the NER task, ACE2005~\citep{doddington2004automatic} and MAVEN~\citep{wang2020maven} for ED, and NYT24~\citep{webnlg2017} for the RE task. {\color{black}The labels in the above datasets follow long-tailed distributions.} We categorize the classes into three splits based on the number of training instances per class, including Few, Medium, and Many, and also report the results on the whole dataset with the Overall setting. We focus more on Mean Recall (MR; \citealt{tang2020unbiased}) and Macro F1 (MF1), two more balanced metrics to measure the performance of long-tailed IE tasks. MR can better reflect the capability in identifying the tail classes, and MF1 can better represent the model's ability for each class, whereas the conventional Micro F1 score highly depends on the head classes and pays less attention to the tail classes. The hyperparameter $\alpha$ in Equation (6) is set as 1 for NER and ED tasks, and $0$ for the RE task. We tune the optimal $\alpha$ on the development sets\footnote{More details are attached in the supplementary materials.}.

\begin{table*}[t]
\centering
\small
\scalebox{0.73}{
\begin{tabular}{lclclclclclclclclc}
\toprule
\multirow{3}{*}{Model}   & \multicolumn{8}{c}{OntoNotes5.0} & & \multicolumn{8}{c}{ATIS} \\
\cline{2-9}\cline{11-18}
                            & \multicolumn{2}{c}{Few} &
                            \multicolumn{2}{c}{Medium}&
                            \multicolumn{2}{c}{Many} &
                            \multicolumn{2}{c}{Overall} & & \multicolumn{2}{c}{Few} &
                            \multicolumn{2}{c}{Medium}&
                            \multicolumn{2}{c}{Many}&
                            \multicolumn{2}{c}{Overall} \\ 
                                               & MR         & MF1       & MR         & MF1         & MR          & MF1     &MR &MF1 &     & MR         & MF1 & MR         & MF1 &MR &MF1        & MR     & MF1       \\
\midrule
BiLSTM~\citep{chiu-nichols-2016}        & 67.5       & \textbf{69.9}   &72.6   &75.3 &88.1 & 85.4 &    76.4    &     76.8             &    & 66.2       & 69.0    &89.8 & 85.9 & 93.0 & 92.2 &    84.2    &   83.1  \\ 
BiLSTM+CRF~\citep{xuezheCRF2016}     &60.7          &     63.6  &65.3 &69.1 & 86.9 & 86.9    &71.6        &  73.5           & &58.1            &60.4  &87.4 &83.5 &93.0 & \textbf{93.5} & 81.0       & 80.2     \\
C-GCN \citep{zhang2018graph}                             &    68.3        & 69.8    &69.1 &72.9   &90.9& 86.6   & 77.3       & 76.8          &  &   63.2         &   65.3 &87.3 &83.2    &91.8 & 89.3    &82.0        & 80.1     \\
Dep-Guided LSTM \citep{jie2019dependency} & 61.8 & 69.3 &70.2 & 73.7 &89.8 & 84.3 &74.1 & 75.8 && 60.6 & 65.4 &94.1 &90.4 &93.2 & 92.9 & 84.6 & 84.3\\
\midrule
%Lifted Loss                                          &            &            &             &              &            &             &        &        &       \\
Focal Loss~\citep{lin2017focal}                                          &     64.1       & 65.5   &69.9 &71.2   &87.7 & 84.7     &    74.2    & 73.9          &         &   48.9         &    49.8  &89.3 &84.6   &91.1 & 89.8    &  78.7      & 76.6        \\ 
%OLTR                                                 &            &            &             &              &            &             &        &        &       \\
cRT~\citep{kang2019decoupling}                                       &     64.1       &    68.5  &73.9 &75.3  &88.0 & 85.2   &  75.0      & 76.1         &     &   68.1         &     71.7   &92.3 &88.0   &92.8 & 92.2  &    85.7    &  84.8      \\
$\tau$ - Normalization~\citep{kang2019decoupling}  &    61.1        &66.7     &72.8 &\textbf{76.4}  &88.0 & 85.7     &73.5        & 75.7            &&     64.8       &  68.0   &89.9 &86.2 & 93.0 & 92.5 & 83.9       & 83.1         \\
LWS~\citep{kang2019decoupling}                                                  &  58.7        &    64.9    &71.6 &76.1   &87.6 & 85.2  &  72.1    & 74.7         &  &66.2& 69.1 &89.9 &85.9 & 93.0 & 92.2 & 84.3 & 83.2  \\
\midrule
%Decoupling~\citep{kang2019decoupling}                                          &            &            &             &              &            &             &        &        &       \\
SGG-TDE \citep{tang2020unbiased}                                                  & 71.9       &    68.8     &77.9 &74.8  & 91.2 & 86.7  & 80.4       &76.7            &         & 67.5      & 67.1  &95.3 &89.4       &93.7 & \textbf{93.5}   &87.1        & 84.5       \\
\midrule
Ours (Glove)                                                 & \textbf{76.7}      & 68.9      &\textbf{83.6} &76.2 & \textbf{92.0} & \textbf{87.6}   &\textbf{83.8}        &\textbf{77.3}           &   & \textbf{71.8}       & \textbf{73.1}       &\textbf{95.6} &\textbf{91.4}  & \textbf{94.3} & \textbf{93.5} &   \textbf{88.6}    & \textbf{87.0}   \\
\midrule
\midrule
BERT \citep{devlin2019bert}                    &    77.7 & 76.5    &81.4 &78.6   & 94.0 & 90.7 & 84.6 & 82.4  &  & 60.7 & 65.9  &97.2 &89.1 & \textbf{93.9} & \textbf{93.5} & 86.1 & 84.0   \\
%Roberta \citep{liu2019roberta}                     &   78.7 & \textbf{79.5}  &84.8 &\textbf{85.1}      & 86.5 & \textbf{85.7}         &  & \textbf{60.2} & 61.1  &93.8 &89.1  & 84.3 & 82.3   \\
%BERT+GCN\citep{wadden2019entity} & 80.2 & 77.7 &81.1 &77.3 & 85.6 & 82.6 & & 53.3 & 56.6 &95.4 &89.6 & 83.0 & 81.3\\
Ours (BERT)                                         & \textbf{80.6} & \textbf{79.1} &\textbf{85.1} &\textbf{80.4}  & \textbf{94.5} & \textbf{91.4} & \textbf{86.7} & \textbf{84.1} &            & \textbf{69.9} & \textbf{71.2}   &\textbf{97.2} &\textbf{91.5}  & 93.5 & 93.1 & \textbf{88.5} & \textbf{86.4}\\
\bottomrule
\end{tabular}
}
\caption{Evaluation results on the OntoNotes5.0 dataset and ATIS datasets for the NER task.}
\label{tab:ontonotes}
%\vspace{-4mm}
\end{table*}

\begin{table*}[t]
\centering
\small
\scalebox{0.73}{
\begin{tabular}{lclclclclclclclclc}
\toprule
\multirow{3}{*}{Model}   & \multicolumn{8}{c}{ACE2005} & & \multicolumn{8}{c}{MAVEN} \\
\cline{2-9}\cline{11-18}
                            & \multicolumn{2}{c}{Few} &
                            \multicolumn{2}{c}{Medium}&
                            \multicolumn{2}{c}{Many} &
                            \multicolumn{2}{c}{Overall} & & \multicolumn{2}{c}{Few} &
                            \multicolumn{2}{c}{Medium}&
                            \multicolumn{2}{c}{Many} &
                            \multicolumn{2}{c}{Overall} \\ 
                                           & MR         & MF1     & MR         & MF1       & MR         & MF1    & MR         & MF1         &  & MR          & MF1     &     MR         & MF1         & MR     & MF1   & MR         & MF1    \\
\midrule
BiLSTM ~\citep{chiu-nichols-2016}                                              & 34.2      &  35.6     & 55.1 & 58.2  &64.9 & 67.0 & 52.3       &54.8    &         & 36.5      &  40.7   &78.3 &79.9 &80.4 & 82.3  & 67.1       &69.5                \\ 
BiLSTM+CRF ~\citep{xuezheCRF2016}     &41.4           &   45.1    &49.8 & 52.2& 70.1 & 70.5    &   51.8     &54.1     &       &43.4           &   46.8    &79.0 &79.8    &82.3 &83.0  &   69.6     &71.1          \\
C-GCN \citep{zhang2018graph}                               &    41.4        &  44.1    &51.2 & 55.8   &66.4 & 71.2  &     52.0   &56.1       &      &   49.7         &51.7     &81.8  & 80.8    &82.6 & 82.1   &    73.1    & 73.0          \\
Dep-Guided LSTM \citep{jie2019dependency} & 42.8 & 41.7  &49.8 & 56.0  &71.1 & 71.6  & 52.4 & 55.8 & & 44.7 & 45.4  &76.5 & 78.2   &75.9 & 78.9  & 67.8 & 69.3 \\
\midrule
Focal Loss~\citep{lin2017focal}                                          &   38.6         & 42.9    &50.7 & 58.8  &74.6 & \textbf{76.0}      &    52.6    &  58.5   &            & 45.4 & 51.5      & 78.6 & 81.3  &85.4 & \textbf{87.2}   & 70.3 & 73.8         \\ 
cRT \citep{kang2019decoupling}                                                 &   44.8         & 47.4     &58.8 & 60.1   &68.8 & 68.5  & 57.6       & 58.9        &     &       49.7     & 55.4     &78.4 & 81.3  &82.1 & 85.0  &    71.0    & 74.6       \\
$\tau$ - Normalization \citep{kang2019decoupling}&     34.3       &     35.6   &50.9 & 53.8  &\textbf{82.7} & 68.3  &    53.3    & 52.5    &        &      21.1      & 26.7   &60.0 & 68.5     &74.4 & 80.0   &  51.0      & 58.4           \\
LWS \citep{kang2019decoupling}                                                 &    34.3        & 35.6     & 61.2 & \textbf{60.2}  &76.8 & 71.7    &58.2        & 56.9   &                 &    33.3        &     38.7  &77.6 &79.7  &81.6 & 81.7   &    65.9    &  68.7       \\
\midrule
SGG-TDE \citep{tang2020unbiased}                             &   34.3    &     33.9      &61.5 & 59.7   &77.4&73.3 &       58.5&   56.5      &            & 39.8       & 36.2     &83.3 & 78.0   &87.8 &85.2  & 71.9       &    67.4         \\
\midrule
Ours (Glove)                                  &    \textbf{47.1}    &     \textbf{49.7}    &\textbf{64.3} & 59.9  &80.5 & 73.3 &        \textbf{63.5}&   \textbf{60.2}        &               & \textbf{60.4}       & \textbf{57.4}       &  \textbf{86.8} & \textbf{82.2}  & \textbf{89.1} & 86.6 &\textbf{79.8}        & \textbf{76.0 }           \\
\midrule
\midrule
BERT\citep{devlin2019bert}      & 47.6 & 48.9    &67.8 & 67.5 &84.5 & 76.8 & 66.5 & 65.1 &&61.1 & 61.8   &86.1 & \textbf{84.6} & \textbf{90.3}& \textbf{89.4} & 79.8 & 78.9 \\
%Roberta \citep{liu2019roberta}    & 47.6 & 47.7 &  73.7 & 72.6      & 70.2 & 68.6 & &\textbf{43.1} & 43.8 &  \textbf{86.3} & \textbf{84.1}& \textbf{75.0} & \textbf{73.6}  \\
%BERT+GCN \citep{wadden2019entity} & 45.2 & 47.5 &\textbf{78.9} & \textbf{77.5} & 73.1 & 71.3 & & 40.9 & 42.0    &86.0 & 83.6  & 74.1 & 72.8  \\
Ours (BERT) & \textbf{61.9} & \textbf{63.2} &\textbf{76.5} & \textbf{76.6} & \textbf{85.3} & \textbf{80.7} & \textbf{74.9} & \textbf{74.4}&& \textbf{61.8} & \textbf{62.7}  &\textbf{86.6} & 84.5 & 90.0& \textbf{89.4}& \textbf{80.2} & \textbf{79.1}  \\ 
\bottomrule
\end{tabular}
}
\caption{Evaluation results on the ACE2005 and MAVEN datasets for event detection.}
\label{tab:ace05}
%\vspace{-4mm}
\end{table*}

\subsection{Baselines}
We categorize the baselines used in our experiments into three groups and outline them as follows.
\noindent
\textbf{Conventional {\color{black}models}} include BiLSTM~\citep{chiu-nichols-2016}, BiLSTM+CRF~\citep{xuezheCRF2016}, C-GCN~ \citep{zhang2018graph}, Dep-Guided LSTM ~\citep{jie2019dependency}, and BERT~\citep{devlin2019bert}. These neural models do not explicitly take the long-tailed issues into consideration.

\noindent
\textbf{Re-weighting/Decoupling models} refer to loss re-weighting approaches including Focal Loss \citep{lin2017focal}, and two-stage decoupled learning methods \citep{kang2019decoupling} that include $\tau$-normalization, classifier retraining (cRT) and learnable weight scaling (LWS).

\noindent
\textbf{Causal model} includes SGG-TDE~\citep{tang2020unbiased}. There are also recent studies based on the deconfounded methodology \citep{tang2020longtailed, yang2020deconfounded} for images, which however seem not applicable to be selected as a causal baseline in our case for text. We ran some of the baseline methods by ourselves since they {\color{black} may have not been reported on NLP datasets.}

%e.g., achieving 10.2 points higher MR against BiLSTM on OntoNotes5.0, and 12.7 points higher MF1 against BiLSTM+CRF on ATIS

\subsection{Task Definitions}

We show the definition of the IE sub-tasks used in our experiments as follows, including named entity recognition (NER), event detection (ED) and relation extraction (RE).

\paragraph{Named Entity Recognition:} NER is a sequence labeling task that seeks to locate and classify named entities in unstructured text into pre-defined categories such as person, location, etc. 

\paragraph{Event Detection:} ED aims to detect the occurrences of predefined events and categorize them as triggers from unstructured text. An event trigger is defined as the words or phase that most clearly expresses an event occurrence. Taking the sentence ``a cameraman died in the Palestine Hotel'' as an example, the word ``died'' is considered as the trigger with a {\color{black}``\textit{Life:Die}''} event. 

\paragraph{Relation Extraction:} The goal of RE is to identify semantic relationships from text, given two or more entities. For example, “Paris is in France” states a “is in” relationship between two entities ``Paris" to ``France". Their relation can be denoted by the triple (Paris, is in, France).

\subsection{Main Results}
\paragraph{NER:} Table~\ref{tab:ontonotes} shows the comparison results on both OntoNotes5.0 and ATIS datasets. Our models perform best or achieve comparable results under most settings, including Few, Medium, Many and Overall. {\color{black}For example, our model} achieves more than 8 points higher MR comparing with the C-GCN model under the Few setting with Glove embeddings on the both of the two benchmarks. The results show the superiority of CFIE in handling the instance-scarce classes for the long-tailed NER. Comparing with a causal baseline SGG-TDE, our model consistently performs better in terms of the two metrics. The results confirm our hypothesis that language structure can help a causal model to better distinguish main effect from the side effect. CFIE also obtains large performance gains with the BERT-based encoder under most of the settings, showing the effectiveness of our approach in mitigating the bias issue with a pre-trained model. {\color{black}It is interesting} that BERT-based models perform worse than Glove-based ones on ATIS. The reason is probably that BERT, which is trained on Wikipedia, may not perform well on a small dataset collected {\color{black}from a very different domain.}    
%Among re-balancing approaches such as Focal Loss, cRT and LWS, $\tau$-Normalization performs best and this aligns with the findings in the previous study \citep{kang2019decoupling} for long-tailed image classification. %It is not supervising that \texttt{BiLSTM+CRF} obtains a slightly better overall F1.

\paragraph{ED:} Table~\ref{tab:ace05} shows comparison results on both of the ACE2005 and MAVEN datasets. Overall, our model significantly outperforms the previous causal baseline SGG-TDE under the Few setting by a large margin, {\color{black}specifically, 12.8 and 15.8 points} higher in terms of MR and MF1 respectively on ACE2005 dataset, 20.6 and {\color{black}21.2 points} higher in terms of the two metrics on MAVEN dataset with Glove embeddings. Meanwhile, our model is able to achieve better or comparable results under other settings, such as Medium and Many. The results further confirm the robustness of our model for tail classes with few training instances available. Our model also performs better or comparable than BERT baselines under Few, Medium, Many and Overall settings, indicating that BERT models still suffer from bias issues on the long-tailed IE tasks. 
%On MAVEN dataset with BERT-based encoders, our CFIE performs worse than Bert under This is probably due to the fact that MAVEN \cite{wang2020maven} has tried to alleviate the unbalanced problem during the data generation.

%RE Table Slim
\newlength{\oldintextsep}
\setlength{\oldintextsep}{\intextsep}
\setlength\intextsep{0pt}
\begin{table}[]%{r}{8.3cm}
\centering
\small
\scalebox{0.82}{
\begin{tabular}{lclclclclc}
\\
\toprule
\multirow{3}{*}{Model}   & \multicolumn{4}{c}{NYT24}  \\
\cline{2-5}
                            & \multicolumn{2}{c}{Few} & \multicolumn{2}{c}{Overall}  \\ 
                                                     & MR         & MF1         & MR          & MF1      \\
\midrule
C-GCN \citep{zhang2018graph}                        &   24.0   &   26.7   &   51.2  &   52.6    \\
\hline
% Focal Loss \citep{lin2017focal}                                          &   52.0    &   48.3    &   62.9    &   61.9    \\
Focal Loss \citep{lin2017focal}                                          &   56.0    &   54.6    &   \textbf{65.7}    &   \textbf{65.5}    \\
cRT \citep{kang2019decoupling}                                                 &   66.0   &   24.2   &   65.6  &   50.5   \\
$\tau$ - Normalization \citep{kang2019decoupling}                        &   40.0   &   40.0   &   53.5  &   54.6    \\
LWS  \citep{kang2019decoupling}                                                &   40.0   &   40.0   &   53.5  &   54.6 \\
\midrule
SGG-TDE \citep{tang2020unbiased}                         &   60.0   &   57.1   &   61.0  &   60.2  \\
\midrule
Ours (Glove)                                         &   \textbf{68.0}   &   \textbf{68.6}   &   65.3  &   63.6    \\
\bottomrule
\end{tabular}
}
\caption{Results on the NYT24 dataset for RE.}
\label{tab:nyt24}
%\vspace{-4mm}
%\lipsum[1]
\end{table}

\begin{figure*}%[!htb]
\setlength{\abovecaptionskip}{0.cm}
\setlength{\belowcaptionskip}{-0.cm}
        \begin{minipage}[t]{0.32\linewidth}
         \centering
         \includegraphics[width=1\textwidth,keepaspectratio]{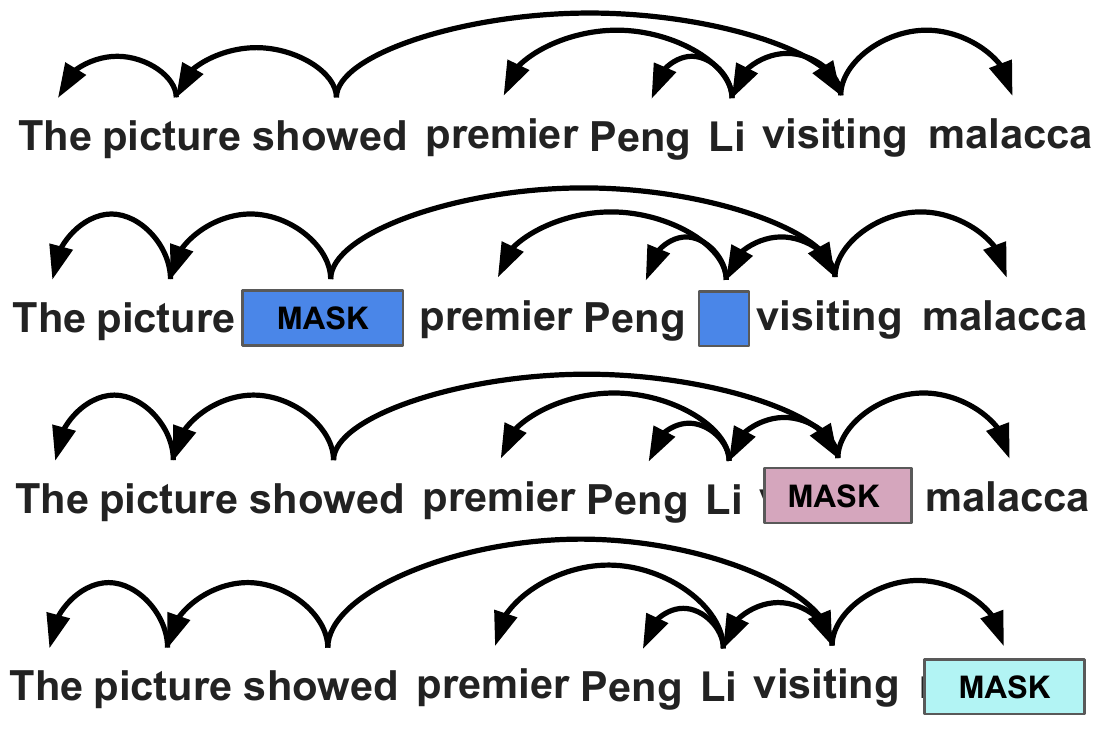}\\
         \caption{Masking operations.}
         \label{fig:ner_deptree}
        \end{minipage}
        \begin{minipage}[t]{0.32\linewidth}
         \centering
         \includegraphics[width=1\textwidth,keepaspectratio]{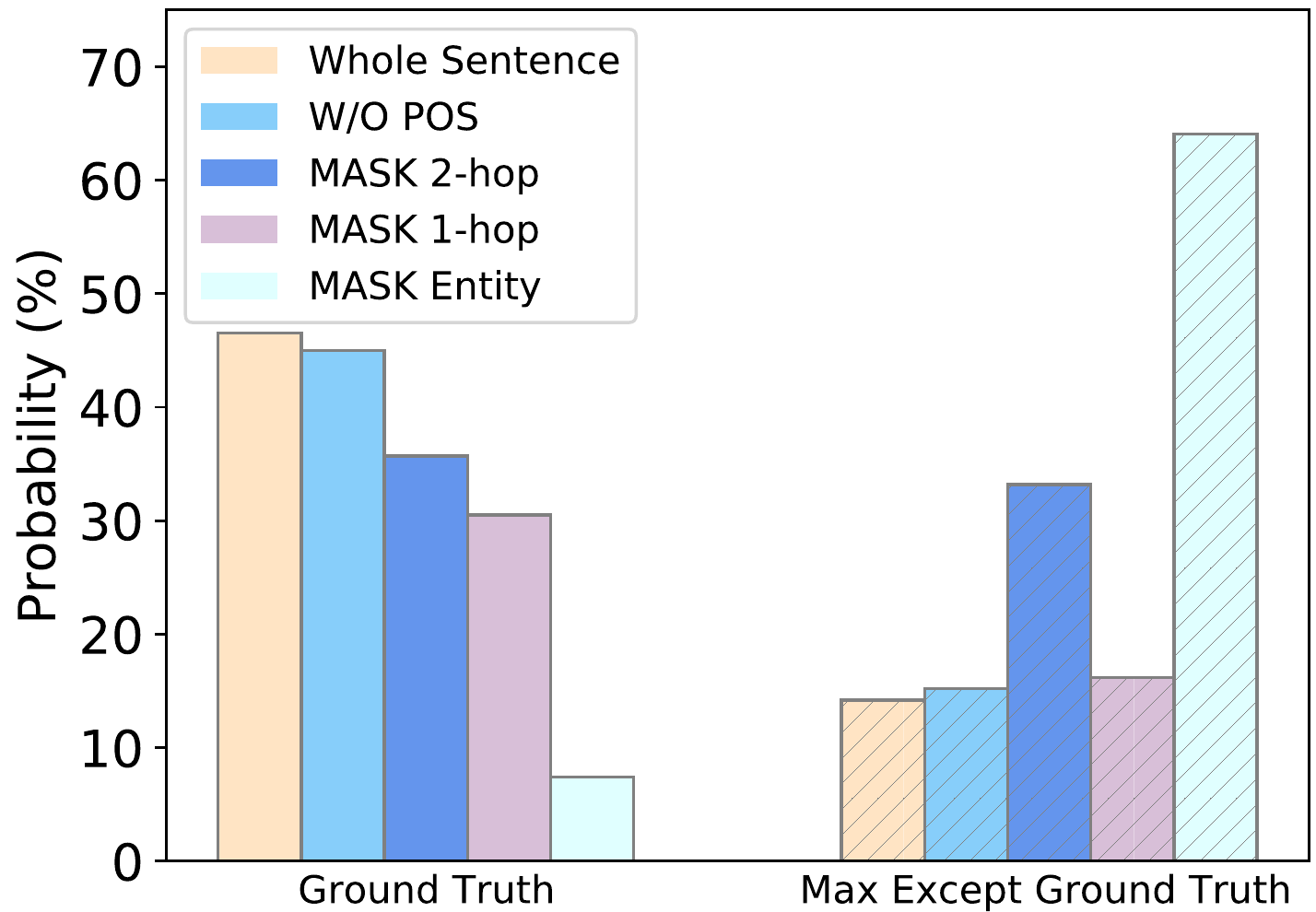}\\
           \caption{Prediction distributions.}
           \label{fig:case_ner}
        \end{minipage}
               \centering
       \begin{minipage}[t]{0.32\linewidth}
         \centering
         \includegraphics[width=1\textwidth,keepaspectratio]{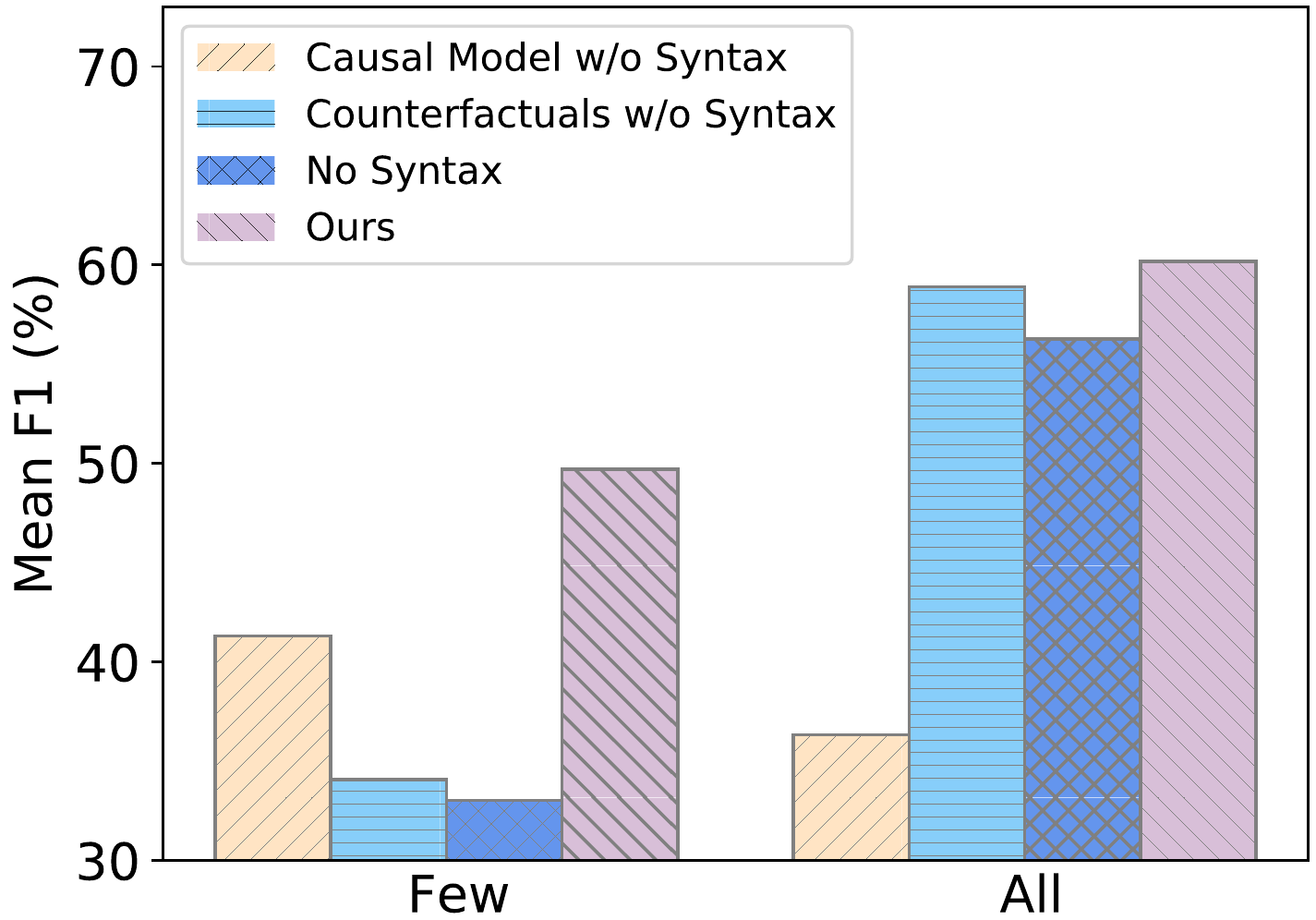}\\
       \caption{Syntax contribution. }
         \label{fig:syntax}
         \end{minipage}  
%         \vspace{-4mm}
\end{figure*}

\paragraph{RE:} As shown in Table~\ref{tab:nyt24}, we further evaluate CFIE on the NYT24 dataset. Our method significantly outperforms all other methods in MR and MF1 for tail classes. The overall performance is also competitive. Although Focal Loss achieves the best overall scores, its ability to handle the classes with very few data points drops significantly, which is the main focus of our work. The results for relation extraction further confirm our hypothesis that the proposed CFIE is able to alleviate spurious correlations caused by unbalanced dataset by learning to distinguish the main effect from the side effect. We also observe that CFIE outperforms the previously proposed SGG-TDE by a large margin for both Few and Overall settings, i.e., 11.5 points and 3.4 points improvement in terms of MF1. This further proves our claim that properly exploring language structure on causal models will boost the performance of IE tasks. %More results are available in the supplementary materials. 

%\texttt{CFIE} is able to alleviate the spurious correlations that may hamper the classifications of the tailed classes. 

\subsection{Discussion}
\label{subsection:discussions}
 %Appendix~\ref{subsection:factor for ed} and \ref{subsection:factor for RE}.
%More analyses about ED and RE are given in the Appendix~\ref{subsection:factor for ed} and \ref{subsection:factor for RE}.

\paragraph{What are the key factors for NER?} 
We have hypothesised that the factors, such as 2-hop and 1-hop context on the dependency tree, the entity itself, and POS feature, may hold the potential to be the key clues for  NER predictions. To evaluate the impact of these factors, we first generate new sequences by {\color{black}masking these factors}. Then we feed the generated sequences to the proposed SCM to obtain the predictions. Figure \ref{fig:ner_deptree} illustrates how we mask the context based on a dependency tree. Figure \ref{fig:case_ner} shows a qualitative example for predicting the NER tag for the entity ``malacca''. It visualizes the variances of the predictions, where the histograms in the left refer to prediction probabilities for the ground truth class, while the histograms in the right are the max predictions except the results of ground truth class. {\color{black}For example, the ``Mask 2-hop" operation with a blue rectangle in Figure \ref{fig:ner_deptree} masks tokens ``showed'' and ``Li'' on the dependency tree, and the corresponding prediction probability distribution is given in Figure \ref{fig:case_ner}, which is expressed as the blue bar}. We observe that masking the entity, i.e., ``malacca'', will lead to the most significant performance drop, indicating that entity itself plays a key role for the NER task. This also inspires us to design a more robust debiasing method as shown in Step $5$ in our framework.  %More analyses about RE are given in the Section $3$ of the supplementary materials.
%Note that a rectangle in Figure \ref{fig:ner_deptree} and a bar in Figure \ref{fig:case_ner} with the same color indicate a masking operation and its corresponding prediction probability.

\paragraph{Does the syntax structure matter?}
To answer this question, we design three baselines including: 1) \texttt{Causal Models w/o Syntax} that doesn't employ dependency trees during the training stage, and only uses it for generating counterfactuals, 2) \texttt{Counterfactuals w/o Syntax} that employs dependency structures for training but {\color{black}uses} a null input as the intervention during the inference state. We refer such a setting from the previous study \citep{tang2020longtailed}, and 3) \texttt{No Syntax} that is the same as the previous work SGG-TDE \citep{tang2020unbiased} which doesn't involve dependency structures in both training and inference stages. As shown in Figure~\ref{fig:syntax}, our model outperforms all three baselines on the ACE2005 dataset under both Few and All settings, demonstrating the effectiveness of dependency structure in improving the 
causal models for the long-tailed IE tasks both in the training and inference stages.%An interesting finding is that dependency-based counterfactual generation brings more gains compare with the gains by applying the structure in the causal model.

%\paragraph{How can we empower a casual inference model by task-specific tree prunning strategies?}
\begin{figure*}%[!htb]
\setlength{\abovecaptionskip}{0.cm}
\setlength{\belowcaptionskip}{-0.cm}
        \begin{minipage}[t]{0.32\linewidth}
         \centering
         \includegraphics[width=1\textwidth,keepaspectratio]{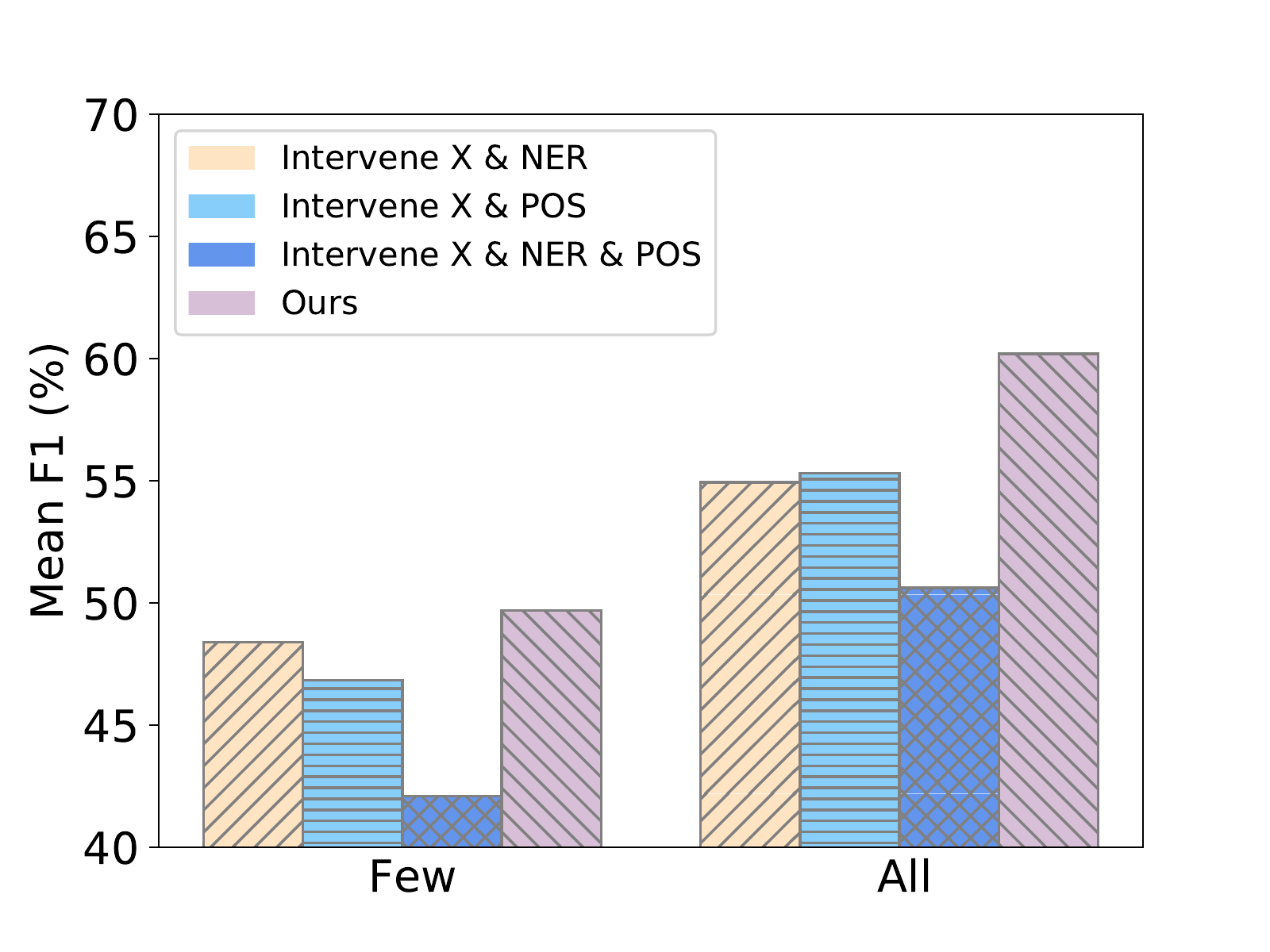}\\
         \caption{Various interventions.}
         \label{fig:diff_intervene_ed}
        \end{minipage}
        \begin{minipage}[t]{0.32\linewidth}
         \centering
         \includegraphics[width=1\textwidth,keepaspectratio]{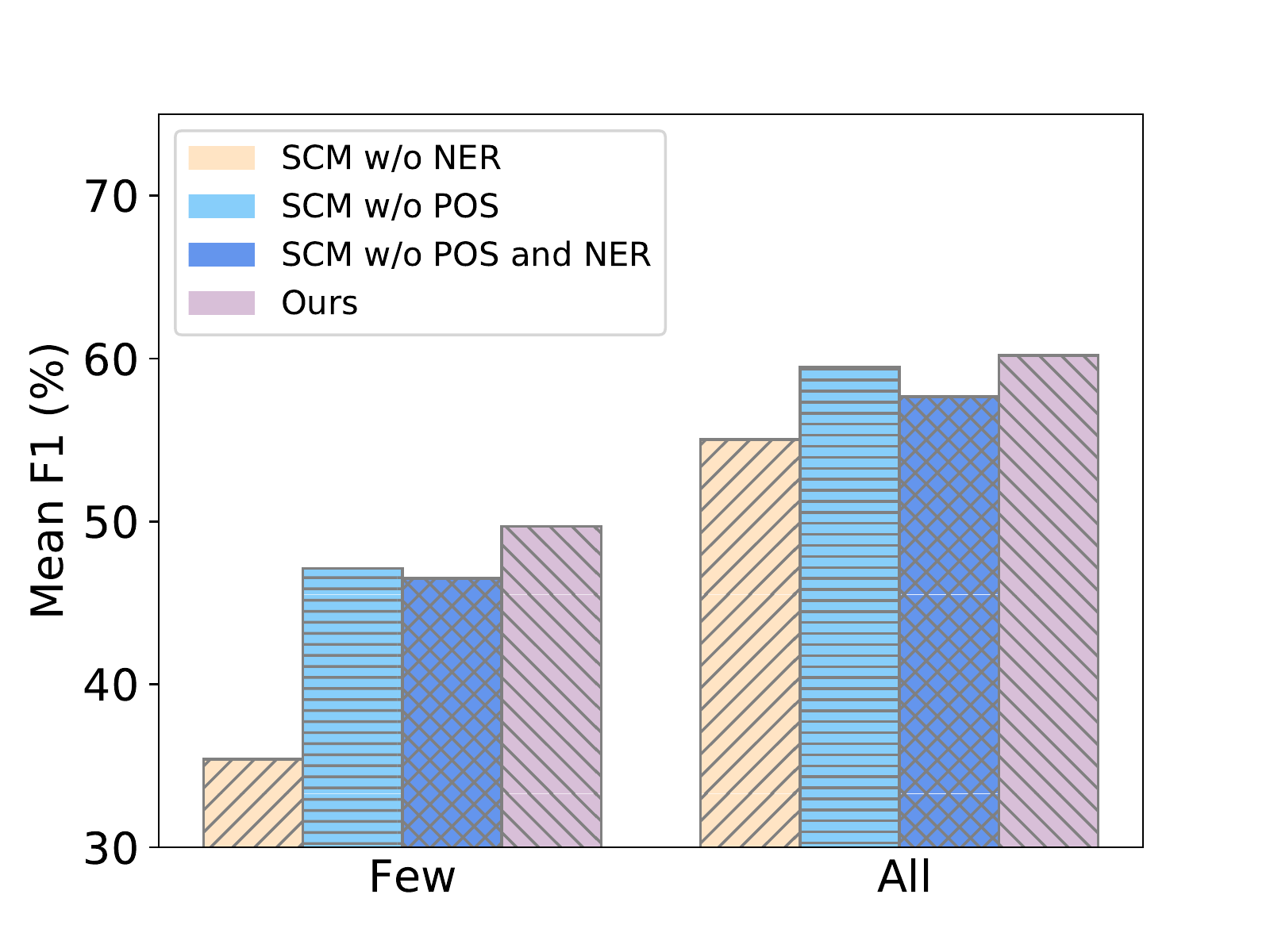}\\
           \caption{Various SCMs.}
           \label{fig:diff_causal_ed}
        \end{minipage}
               \centering
       \begin{minipage}[t]{0.3\linewidth}
         \centering
         \includegraphics[width=1\textwidth,keepaspectratio]{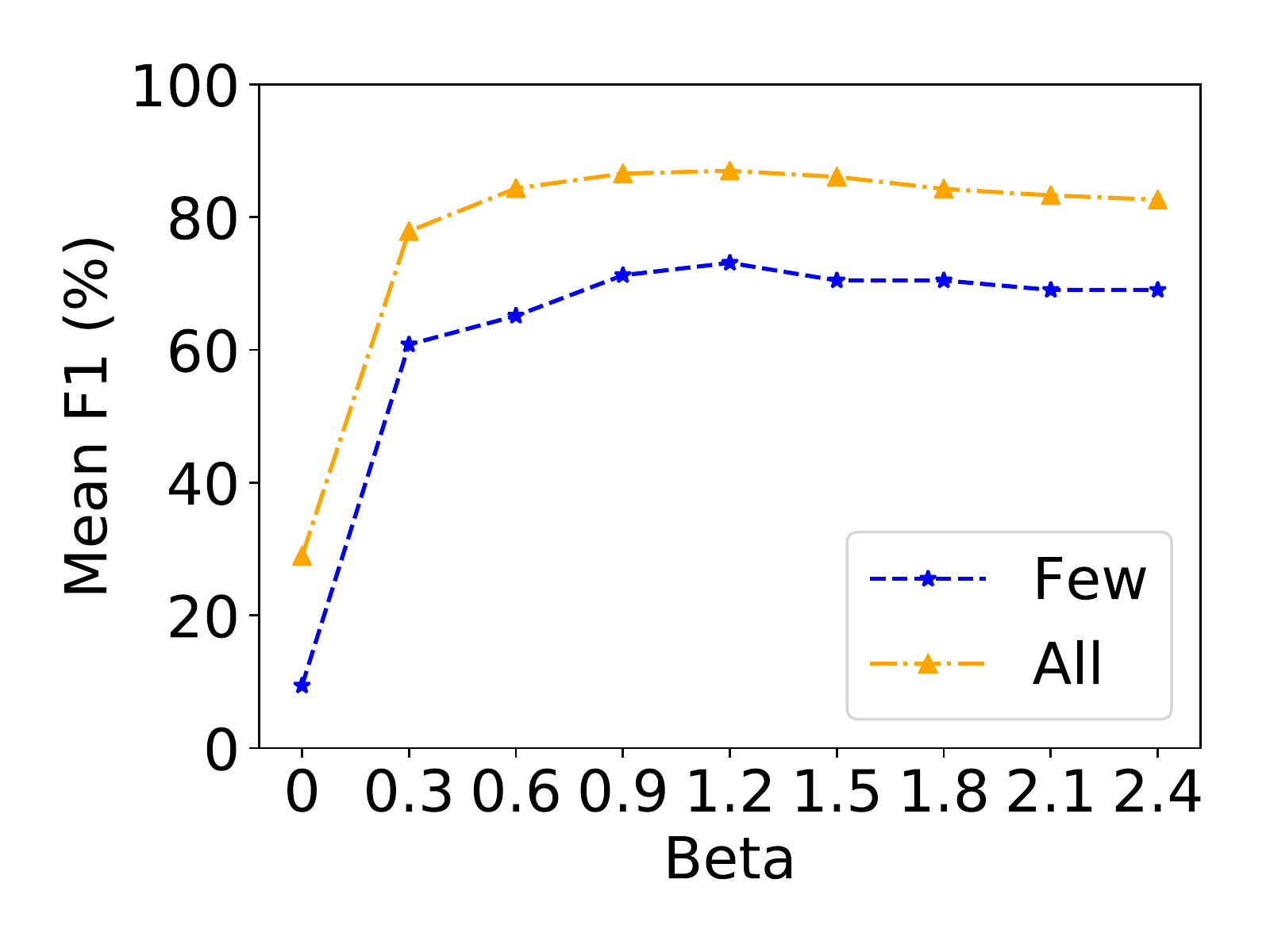}\\
       \caption{Various $\beta$. }
         \label{fig:alpha_atis}
         \end{minipage}  
%         \vspace{-4mm}
\end{figure*}

\iffalse
\paragraph{How can we make good use of dependency structure?} To answer this question, we present three tree pruning mechanisms under two graph aggregation settings, i.e., \texttt{Prune with DGLSTM} based on Dep-Guided LSTM \cite{jie2019dependency} and \texttt{Prune with C-GCN} based on GCN \cite{Kipf2016SemiSupervisedCW} as described in Section 2.1. The three pruning strategies include 1) \texttt{CFIE Mask 1-hop} which masks the tokens that directly connect to the targeting token along a dependency tree, 2) \texttt{CFIE Mask token} which directly masks the targeting token, 3) \texttt{CFIE Mask token\&1-hop} which masks both the targeting token and its 1-hop neighbours in the dependency tree.
Figure \ref{fig:prunning-dep} and Figure \ref{fig:pruning-gcn} depict the results on OntoNotes5.0 dataset. We observe that masking 1-hop neighbours in the syntactic tree achieves the best performance among three strategies, indicating that an entity itself is more important in NER sequence labeling. By comparing the two graph aggregation method, we draw a conclusion that \texttt{Prune with DGLSTM} can make better use of dependency structures.
%The intuition behind the above operations is that tokens selected by a dependency tree may contain some clues for the classification, and our aim here is to study the impacts of different tree prunning strategy on address in instance-scarce 
%One interesting question is which variable(s) we should intervene to explore the SCM.
\fi

\paragraph{How {\color{black}do} various interventions and SCMs affect performance?} We study this question on ACE2005 dataset for ED task. We design three interventional methods including 1) \texttt{Intervene X \& NER}, 2) \texttt{Intervene X \& POS}, 3) \texttt{Intervene X \& NER \& POS} . Figure~\ref{fig:diff_intervene_ed} shows that  introducing interventions solely on $X$ is able to achieve the best performance under both Few and All settings. We also introduce three variants of our proposed SCM : 1) \texttt{SCM w/o NER}, 2) \texttt{SCM w/o POS}, 3) \texttt{SCM w/o NER and POS}. Figure~\ref{fig:diff_causal_ed} shows that {\color{black}removing} the NER node will significantly decrease the ED performance, especially over the Few setting. The results prove the superiority of our proposed SCM that explicitly involves linguistic features to calculate main effect. %More analyses for NER task are given in Appendix~\ref{subsection:SCM for NER}. 

\begin{figure*}
    \centering
    \includegraphics[scale=0.63]{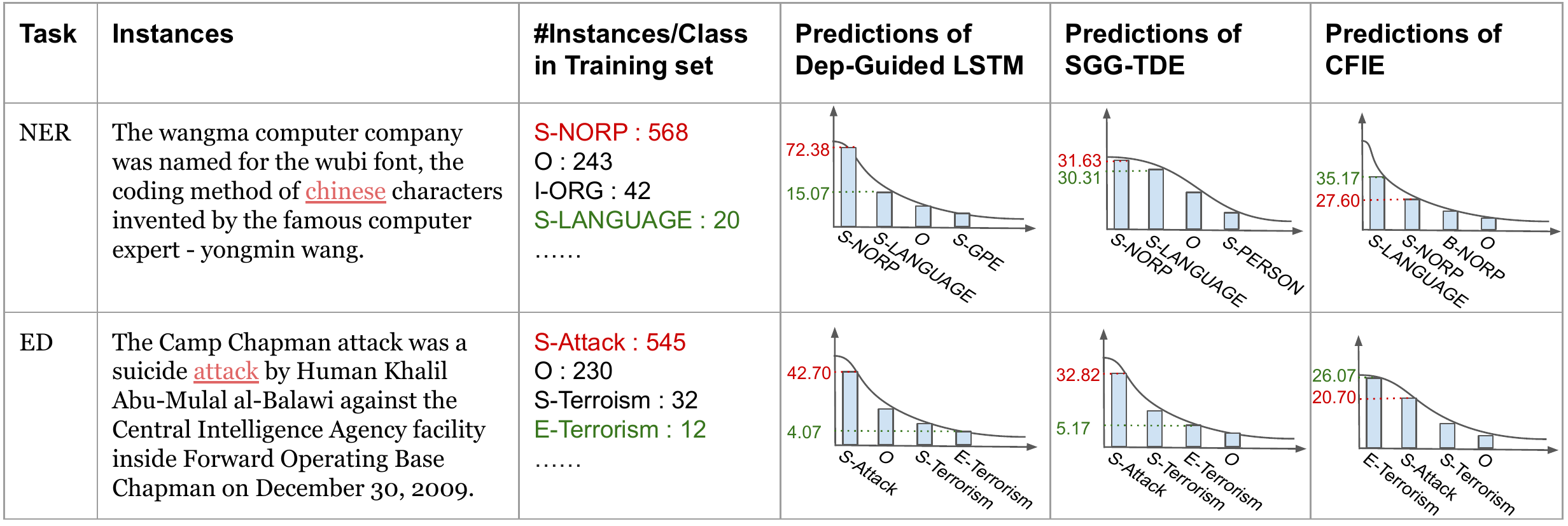}
%    \vspace{-1mm}
    \caption{Two cases selected from OntoNote5.0 and MAVEN for NER and ED tasks respectively, with unbalanced distributions for the targeting entity and event trigger. The two baseline models Dep-Guided LSTM and SGG-TDE tend to predict incorrect results caused by the spurious correlations, while our proposed CFIE model is able to yield better predictions.}
    \label{fig:casestudy}
%    \vspace{-4mm}
\end{figure*}

\paragraph{How {\color{black}does }the hyper-parameter $\beta$ impact the performance?} To evaluate the impact of $\beta$ on the performance, we tuned the parameter on four datasets including OntoNotes5.0, ATIS, ACE2005, and MAVEN. As shown in Figure~\ref{fig:alpha_atis}, when increasing $\beta$ from $0$ to $2.4$ on ATIS dataset, the F1 scores increase  dramatically then decrease slowly. The F1 scores reach the peak when $\beta$ is 1.2. As the value of $\beta$ represents the importance of entity for classifications, we therefore draw a conclusion that, for NER task, an entity plays a relatively more important role than the context \cite{zeng2020counterfactual}. We observe that the performance significantly drops when $\beta$ is 0. {\color{black}This suggests that directly applying previous causal approach \cite{tang2020unbiased} may not yield good performance.} The result further confirms the effectiveness of Step $5$ in CFIE. %More experimental results on the other datasets are given inthe supplementary materials. 

\subsection{Case Study}
Figure~\ref{fig:casestudy} shows two cases to visualize the predictions of baseline models and our CFIE model for long-tailed NER and ED, respectively. We use {\color{black}the} ``BIOES'' tagging scheme for both cases and choose Dep-Guided LSTM \cite{jie2019dependency} and SGG-TDE \cite{tang2020unbiased} as baselines. In the first case for NER, the baseline {\color{black}assigns} ``chinese'' with the label S-NORP, which indicates ``nationalities or religious or political groups'', while the corrected annotation is ``S-LANGUAGE''. This is caused by the spurious correlations between ``chinese'' and S-NORP learned from unbalanced data. For this case, there are $568$ and $20$ training instances for S-NORP and S-LANGUAGE, respectively. The numbers of training instances for each type are indicated in the third column of Figure~\ref{fig:casestudy}. The numbers in the $4$-th to $6$-th columns indicate the probability of the token ``chinese'' predicted as a certain label. For example, in the $6$-th column “Predictions of CFIE”, the prediction probability is $35.17\%$ for the label S-LANGUAGE. In the second case for ED, we demonstrate a similar issue for the trigger word ``attack'', and compare it with the two baselines. For both cases, previous SGG-TDE outputs relatively unbiased predictions compared with Dep-Guided LSTM, although the predictions are also incorrect. Our CFIE model can obtain correct results for both instances, showing the effectiveness of our novel debiasing approach. {\color{black}Compared to CFIE, the inferior performance of SGG-TDE is due to ignoring the importance of entity (trigger) for the NER and ED tasks.} %  %as described in Equation (\ref{eq:tde}) and (\ref{eq:debias}).       
\section{Related Work}
\noindent
\textbf{Long-tailed IE:} RE \citep{Zeng2014RelationCV, Peng2017CrossSentenceNR, Quirk2017DistantSF,song2018n,lin2019learning,peng2020learning,nan2020reasoning,qu2020few,guo2020learning,zhang2021document,zheng2021prgc,zhang2020openue,ye2021contrastive,bai2021semantic,nandialog2021}, NER \citep{lampleetal2016, chiu-nichols-2016, xu2021better}, and ED \citep{nguyengrishman2015event, huang-etal-2018} are mainstream IE tasks in NLP. For the long-tailed IE, recent models \citep{lei-etal-2018, zhangetal2019long} leverage external rules or transfer knowledge from data-rich classes to the tail classes. 
Plenty of re-balancing models are also proposed, including re-sampling strategies \citep{mahajan2018exploring, wang2020devil} that aim to alleviate statistical bias from head classes, and re-weighting approaches \citep{milletari2016v, lin2017focal} which assign balanced weights to the losses of training samples from each class to boost the discriminability via robust classifier decision boundaries. Another line is decoupling approaches \citep{kang2019decoupling} that decouple the representation learning and the classifier by direct re-sampling. Different from the above works that are based on conventional approaches, we tackle the long-tailed IE problem from the perspective of causal inference.

\noindent
\textbf{Causal Inference:} Causal inference \citep{pearl2016causal,rubin2019essential} has been applied in many areas, including visual tasks \citep{tang2020unbiased, abbasnejad2020counterfactual, niu2020counterfactual, yang2020deconfounded, zhang2020causal, yue2020interventional,yang2021deconfounded,nan2021interventional}, model robustness and stable learning \cite{srivastava2020robustness,zhang2020causal,shen2020stable, yu2020label,dong2020counterfactual}, generation \cite{wu-etal-2020-de}, language understanding \cite{feng2021empowering}, and recommendation systems \cite{jesson2020identifying,feng2021should,zhang2021causal,wei2021model,wang2021clicks,tan2021counterfactual,wang2021counterfactual,ding2021causal}. Works most related to ours are ~\citep{zeng2020counterfactual, wang2020robustness} that generates counterfactuals for weakly-supervised NER and text classifications, respectively. Our method is remotely related to \cite{tang2020unbiased} proposed for image classifications. The key differences between our methods and previous ones: 1) counterfactuals in our method are generated by a task-specific pruned dependency structure on various IE tasks. While in previous works, counterfactuals are generated by replacing the target entity with another entity or their antonyms \cite{zeng2020counterfactual, wang2020robustness}, or simply masking the targeting objects in an image \cite{tang2020unbiased}. These method do not consider the complex language structure that has been proven useful for IE tasks. 2) compared with previous method SGG-TDE \cite{tang2020unbiased}, our inference mechanism is more robust for various IE tasks, simultaneously mitigating the spurious correlations and strengthening salient context.

\iffalse
\noindent
\textbf{Model Interpretation:} There have been plenty of studies \citep{molnar2020interpretable} about traditional model interpretation applied in various applications, such as text and image classification \citep{ribeiro2016should,ebrahimi2018hotflip}, question answering~\citep{feng2018pathologies,ribeiro2018semantically}, and machine translation~\citep{doshi2017towards}.  LIME~\citep{ribeiro2016should} was proposed to select a set of instances to explain the predictions. The input reduction method \citep{feng2018pathologies} finds out the key features and uses very few words to obtain the same prediction. Unlike the two methods, the selections in our CFIE model are based on a syntactic tree. Some recent studies \citep{ribeiro2018semantically,kaushik2019learning} use data augmentation with extra training signal. Our CFIE model is orthogonal to these methods for training purpose. We generate counterfactuals during the inference stage.
\fi

\section{Concluding Remarks}
This paper presents CFIE, a novel framework {\color{black}for} tackling the long-tailed IE issues in the view of causal inference. Extensive experiments on three popular IE tasks, named entity recognition, event detection, and relation extraction, show the effectiveness of our method. Our CFIE model provides a new perspective on tackling spurious correlations by exploring language structures based on structured causal models. We believe that our models {\color{black}may also find applications} in other NLP tasks that suffer from spurious correlation issues caused by unbalanced data distributions. Our future work includes developing more powerful causal models for the long-tailed distribution problems using the task-specific language structures learned from the data. {\color{black}We are also interested in addressing the spurious correlations in various vision and language tasks}  \cite{nan2021interventional,IVRD2021MM,xu2021sutd,fan20dialog,liu2021context,zhang2021video,zhang2021natural,chen2020scanrefer,chen2021scan2caps}. %We leave it as our future study. %Future research directions include extending the proposed framework to more challenging document-level IE tasks, where the text structure may need to be learned from the data.

\section*{Acknowledgments}
%\vspace{-1mm}
We would like to thank the anonymous reviewers for their thoughtful and constructive comments. 
%This research is supported by Ministry of Education, Singapore, under its Academic Research Fund (AcRF) Tier 2 Programme (MOE AcRF Tier 2 Award No: MOE2017-T2-1-156). Any opinions, findings and conclusions or recommendations expressed in this material are those of the authors and do not reflect the views of the Ministry of Education, Singapore.
This work is done when Jiaqi Zeng was working as a research intern at SUTD and when Rui Qiao was working as a research assistant at SUTD.
This project/research is supported by the National Research Foundation, Singapore under its AI Singapore Programme (AISG Award No: AISG-RP-2019-012).

\bibliography{anthology,custom}
\bibliographystyle{acl_natbib}

%\clearpage
%\appendix

%\input{appendix}
%\section{Example Appendix}
%\label{sec:appendix}

%This is an appendix.

\end{document}